\documentclass[lettersize,journal]{IEEEtran}
\usepackage{amsmath,amsfonts}
\usepackage{algorithmic}
\usepackage{array}
\usepackage[caption=false,font=normalsize,labelfont=sf,textfont=sf]{subfig}
\usepackage{textcomp}
\usepackage{stfloats}
\usepackage{url}
\usepackage{verbatim}
\usepackage{graphicx}
\usepackage{array}
\usepackage{booktabs}
\usepackage{float} 
\usepackage{multirow}
\def\BibTeX{{\rm B\kern-.05em{\sc i\kern-.025em b}\kern-.08em
    T\kern-.1667em\lower.7ex\hbox{E}\kern-.125emX}}
\usepackage{balance}
\begin{document}
\title{SyLink Hand: A Synergy-Inspired Linkage-Driven Anthropomorphic Hand for Human-Like Dexterity
}
\author{Hao Wu$^{1}$, Yanzhe Wang$^{2}$, Yu Feng$^{1}$, Yitong Li$^{1}$, Jingxiang Guo$^{1}$, Jian Liu$^{1}$, and \\Jianshu Zhou$^{1}$,~\IEEEmembership{Member,~IEEE,~ASME}

\thanks{1.\,Department of Mechanical Engineering, National University of Singapore, Singapore. 2.\,Department of Mechanical Engineering, Zhejiang University, Hangzhou, China. Corresponding author: Jianshu Zhou (jianshuzhou@nus.edu.sg)
}}

\markboth{}%
{SyLink Hand: A Synergy-Inspired Linkage-Driven Anthropomorphic Hand for Human-Like Dexterity}

\maketitle

\begin{abstract}
Designing anthropomorphic robotic hands that balance functional dexterity with mechanical simplicity remains a significant challenge. Inspired by human hand synergies, this paper presents the SyLink Hand, an anthropomorphic dexterous hand that integrates biomechanical synergy principles with linkage-driven transmission mechanisms to achieve a high degree of anthropomorphism in appearance, kinematics, and functionality within a compact and cost-effective architecture. Biomechanical analysis of natural hand motions using motion capture gloves reveals strong kinematic correlations among hand joints, providing the basis for a simplified yet functional degree-of-freedom (DOF) configuration. Guided by these synergistic characteristics, optimized linkage mechanisms are employed to coordinate multiple joint motions and reproduce natural finger trajectories. A novel spherical four-bar linkage is further proposed to achieve decoupled flexion/extension (Flex/Ext) and abduction/adduction (Abd/Add) at the metacarpophalangeal joint within a compact form factor. The resulting prototype integrates 19 joints driven by 11 actuators, with a total mass of 520g and a manufacturing cost of approximately USD 400. Experimental evaluations demonstrate its human-like kinematic performance, high load-bearing capability, and versatile grasping and manipulation skills. These results validate that the synergy-inspired, linkage-based design effectively balances anthropomorphism, mechanical simplicity, and functional versatility, highlighting its potential for practical deployment in dexterity-demanding robotic applications.

\end{abstract}


\section{Introduction}

\IEEEPARstart{T}{he} human hand is widely regarded as one of the most intricate parts of the human body, serving as a critical medium for object manipulation and environmental interaction. Its high degree of freedom, precise muscle-joint coordination, and multimodal sensory feedback endow it with remarkable flexibility and adaptability in addressing complex grasping and manipulation tasks. From multi-fingered \cite{guo2025enabling,wang2025flexible,chen2021design,zhou2024dexterous} grippers to anthropomorphic hands \cite{kim2021integrated,liu2008multisensory,su2024tit,ren2023novel,zhou2019soft}, classic design approaches have consistently endeavored to replicate the features, dexterity, and functionality of human hands with sophisticated mechanisms integrating more actuators and sensors. However, this design paradigm often leads to increased structural complexity, high cost, and control difficulties, resulting in limited applications in real-world scenarios. Therefore, significant challenges remain in seeking ideal functional anthropomorphism and compact transmission structures for robotic hands.

\begin{figure}[!t]
    \centering
    \includegraphics[width=1\linewidth]{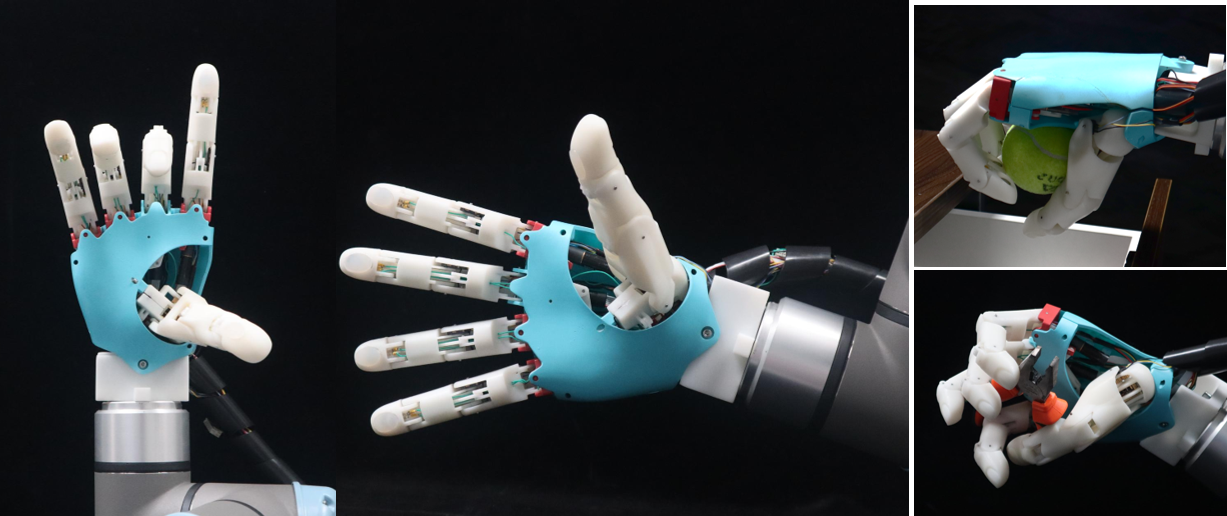}
    \caption{The SyLink Hand prototype with human-like appearance and dexterous grasping capability.}
    \label{FigureLabel1}
\end{figure}

The concept of hand synergy from neuroscience \cite{ingram2008statistics,gracia2020sharing} provides a promising approach to reducing this complexity while preserving main functionalities. Extensive related studies have been conducted to deepen our understanding of the hand's grasping strategies, synergistic characteristics, and methodologies for translating this knowledge into robotic hand design. For instance, Xiong et al. \cite{xiong2016design} applied principal component analysis (PCA) to motion data obtained using the sensor glove and proposed a universal design theory for prosthetic hands. By leveraging the identified key biomechanical synergies, the prototype X-hand is actuated by only four motors while demonstrating the capability to perform 91\% of daily human grasping activities. Similarly, Sun \cite{sun2021design} presented a design principle for a robotic hand powered by only two actuators, enabling it to replicate a wide range of human grasping gestures through independent thumb movement and synergistic finger actions. Additionally, this concept demonstrates strong practicality in the design of task-specific robotic hands. Inspired by efficient swing mechanics, Chang \cite{chang2021anthropomorphic} presented an anthropomorphic prosthetic hand designed specifically for sports activities. It incorporates human anatomical features and tendon routing to enhance swing efficiency, enabling amputees to achieve a 19\% increase in swing speed. Karnati \cite{karnati2012bioinspired} investigated human finger synergies during screwing and unscrewing motions, and subsequently developed sinusoidal joint trajectories to enable a robotic hand to perform analogous tasks with a single input. Although the complexity reduction brought by the synergistic organization in these designs enables the replication of most human grasping capabilities or specific tasks, the inherent limitation in degrees of freedom restricts their application in more complex manipulation and general operation tasks.

The integrity and versatility of robotic hands largely depend on the design and implementation of their joint mechanisms. In particular, we focus on the methodologies for implementing coupled joints and 2-DOF base joints. In the realm of coupled joint motion, Yoon \cite{yoon2017underactuated} presented an underactuated finger mechanism that utilizes contractible slider-cranks and stackable four-bar linkages to achieve adaptive and efficient grasping with reduced actuation requirements. Xu \cite{xu2013design} proposed a mechanical design with flexible shafts connected to worm gears and spur gear trains to drive the finger joints. This arrangement enables joint coupling, ensuring coordinated motion between paired joints and synchronized abduction of multiple fingers, thus achieving the desired postural synergies. In contrast, the LISA Hand \cite{jin2012lisa} achieves underactuation through a linkage-based indirect self-adaptive mechanism, harnessing the inverse forces from grasped objects to drive subsequent joint actions without additional motors.
Meanwhile, the realization of 2-DOF in the base joint of the fingers is crucial for enhancing manipulation capabilities. The DLR/HIT Hand II \cite{liu2008multisensory} employs a differential gear mechanism, where two actuators share the load and collectively generate 2-DOF motion through synchronized and counter-rotational movements of the bevel gears. Similarly, the ILDA Hand \cite{kim2021integrated} achieves dexterous control of 2-DOF movements through a hybrid serial-parallel linkage mechanism that leverages two linear motions. Although these structures enable independent movement in two directions, they necessitate the simultaneous operation of two motors, inherently limiting the flexibility in configuring the degrees of freedom. To address this limitation, Li \cite{li2023linkage} proposed a differential linkage mechanism enabling fully decoupled actuation, though its large spatial footprint limits its practicality in space-constrained applications.

% By contrast, higher degrees of freedom necessitate a more compact structural configuration, posing challenges to the design of dexterous hands. Research into dexterous hands with high/full degrees of freedom is underway, utilizing mechanisms such as tendons, modular fingers, and linkages to achieve sophisticated manipulation capabilities. Chen's research has contributed to the development of the ILDA hand, a robotic hand that employs linkage-driven mechanisms to achieve a high degree of freedom (15-DOF), allowing for complex manipulation tasks and a human-like grasp \cite{kim2021integrated}. Furthermore, the Shadow Hand is a sophisticated robotic hand with 24 degrees of freedom, designed to closely mimic the human hand's kinematics and dexterity \cite{walkler2004developments}. Puhlmann presented the RBO Hand 3, a highly capable and versatile anthropomorphic soft hand based on pneumatic actuation. The RBO Hand 3 possesses 16 independent degrees of actuation, facilitating dexterous manipulation and human-like grasping strategies with low-cost and easily accessible materials \cite{puhlmann2022rbo}. The KITECH-Hand's modular architecture and distinctive metacarpophalangeal joint mechanism augment its kinematic capabilities and simplified mechanical design, thereby achieving enhanced dexterity over conventional robotic hands \cite{lee2016kitech}. Despite their high degree of freedom, these dexterous hands exhibit complexity in control and bulkiness in structure. And actuation methods such as tendons and pneumatics significantly reduce motion precision.

Motivated by the unsolved issues in the aforementioned areas, in this work, we attempt to investigate biomechanical mechanisms of the human hand synergy through the analysis of the correlation relationship in natural hand movements, and present a synergy-based strategy for actuation configuration without sacrificing high-degree-of-freedom functionality. Moreover, we propose the corresponding linkage mechanisms to implement these features in the robotic hand. Specifically, diverse configurations of planar linkage mechanisms are employed to enable joint coupling and, through optimized positions and lengths, effectively replicate the natural motion trajectories of the human hand. For the 2-DOF joint, traditional implementation methods such as the differential gear mechanism necessitate the concurrent operation of two motors to control two degrees of freedom, which is incompatible with our requirement for synchronized lateral motion. Consequently, a spherical four-bar mechanism has been designed and validated to replace this mechanism. The final prototype features a human-like appearance and movement, as shown in Fig. \ref{FigureLabel1}, with a compact and efficient structure integrated within the hand. Experimental results demonstrate that the proposed hand exhibits human-like motion dexterity, robust force generation and load-bearing performance, and sensitive force perception, thereby enabling versatile object grasping, in-hand manipulation, and task-oriented tool operation. The primary contributions of this work are summarized as follows:

\begin{figure}[!t]
    \centering
    \includegraphics[width=1\linewidth]{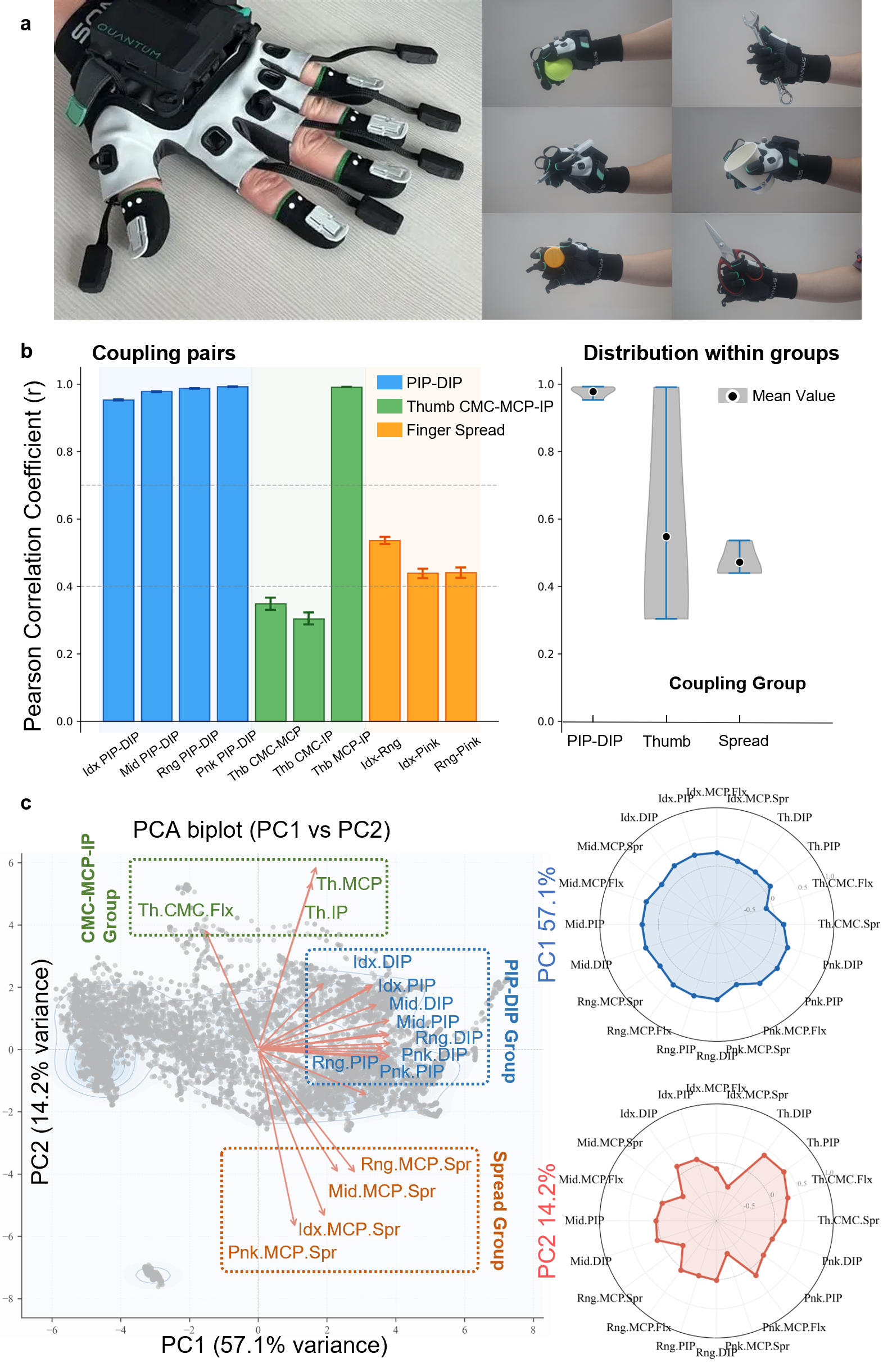}
    \caption{Human hand motion data acquisition and synergy analysis. (a) Experimental setup for collecting joint angles across representative grasp types. (b) Pearson correlation coefficients of inter-joint synergies across PIP–DIP, thumb CMC–MCP–IP, and finger spread joint groups. (c) PCA visualization of hand joint synergies.}
    \label{FigureLabel2}
\end{figure}

\begin{itemize}
\item[(1)] The biomechanical characteristics of human hand synergies are analyzed and extracted to guide the mechanical design of the robotic hand. Based on these characteristics, a simplified DOF  distribution and hand configuration are formulated to minimize structural complexity while preserving essential dexterous functionalities.

\item[(2)] Linkage-based coupling mechanisms are employed to realize the coordinated motions of the proximal interphalangeal (PIP) and distal interphalangeal (DIP) joints of the four fingers, the carpometacarpal (CMC), metacarpophalangeal (MCP), and interphalangeal (IP) joints of the thumb, as well as lateral motions of the index, ring, and little fingers. The linkage parameters are further optimized to reproduce natural human hand motion patterns.

\item[(3)] A spatial spherical linkage mechanism is proposed for the actuation of 2-DOF joints. The mechanism achieves effective decoupling of Flex/Ext and Abd/Add motions, enabling independent control of each degree of freedom within a constrained space while supporting synergistic underactuated lateral motion across multiple fingers.

\item[(4)] An integrated anthropomorphic hand architecture, including the overall appearance, DOF configuration, and structural layout, is developed to closely emulate the morphology and functionality of the human hand, achieving a high degree of anthropomorphism while maintaining mechanical simplicity and transmission efficiency.

\end{itemize}

\section{Biomimetic Analysis}

The human hand is a highly sophisticated musculoskeletal system with 22 joints and nearly 38 muscles, along with an intricate network of tendons and thousands of sensory receptors \cite{schwarz1955anatomy}. This complex structure enables remarkable dexterity and functionality but also makes it significantly challenging to replicate in robotic hands. Therefore, simplifications and trade-offs are necessary to strike a balance between complexity and functionality, enabling the development of practical actuation strategies and optimized degrees-of-freedom configurations for robotic hands.

% Here, we use the CyberGlove II to record the joints angle of the human hand during grasping objects with different shapes. Three healthy right-hand subjects wearing the CyberGlove are instructed to grasp each object shown in Fig. 2 for three times in appropriate grasp patterns. The initial posture of each grasp is with fully extended fingers and the abduction between the thumb and index finger is in the comfortable status (see Fig. 2). Each grasping movement is averagely performed in 3 s including one second for the start and one second for the end of the grasp to ensure the zero velocity at the two ends. The involved grasp patterns follow the grasp taxonomy proposed by Cutkosky [33]. In our experiment, each object is grasped in different feasible types as many as possible. Finally, there are 322 grasping movements performed in total.

To investigate the kinematic mechanisms and synergistic features of the human hand, an experimental protocol was established to collect a dataset of human hand movements, which served as the basis for the design of the proposed anthropomorphic hand. In this experiment, we used the Manus Quantum Mocap Metagloves to capture the joint angles of the human hand during grasping tasks involving objects of various shapes and sizes, performed in the subject's natural manner (shown in Fig.~\ref{FigureLabel2}a). Participants were instructed to grasp and manipulate target objects in their habitual manner, performing each task three times to minimize random errors. Throughout the grasping actions, the Metagloves recorded angular measurements of 20 joints. All measured joint angles formulate the human hand movement dataset $Q$ for the grasping activities.

After obtaining the human hand movement dataset, we can quantitatively analyze the synergistic principles between the joints and the digits. PCA is frequently employed to extract the fundamental synergies in human hand gestures, aiding in the analysis and optimization of the configuration of DOFs in dexterous hands \cite{xiong2016design,karnati2012bioinspired,laffranchi2020hannes}. Here, the correlation coefficient is calculated according to the following formula, resulting in a 20 × 20 correlation matrix $R$:
\begin{equation}
\mathrm{Cov}(Q_i,Q_j)
=
\frac{(Q_i-\bar{Q}_i)^T(Q_j-\bar{Q}_j)}
{n-1}
\end{equation}
\begin{equation}
\sigma_i
=
\sqrt{
\frac{(Q_i-\bar{Q}_i)^T(Q_i-\bar{Q}_i)}
{n-1}
}
\end{equation}
\begin{equation}
r_{i,j}
=
\frac{\mathrm{Cov}(Q_i,Q_j)}
{\sigma_i \sigma_j}
\end{equation}

The resulting correlation matrix $R$ provides a comprehensive representation of the synergistic relationships among the hand joints. Since the objective of this analysis is to identify joint groups suitable for mechanical coupling, only correlations between anatomically adjacent joints are evaluated. Correlations involving non-adjacent joints are disregarded because realizing direct coupling between them would require complex transmission mechanisms and is generally infeasible in a compact anthropomorphic hand. Strong correlations are observed among the PIP and DIP joints of the four fingers, the CMC, MCP, and IP joints of the thumb, as well as the joints responsible for lateral motion of the four fingers. The results are shown in Fig.~\ref{FigureLabel2}b. The PCA results further confirm these coupling patterns. Based on the PC1‑PC2 biplot presented in Fig.~\ref{FigureLabel2}c, vectors within each coupling group point in consistent directions, whereas those from distinct groups are approximately orthogonal. This observation demonstrates strong intra‑group coordination and inter‑group independence. These coupling patterns are consistent with natural hand kinematics observed in daily activities, where such joints frequently operate in a coordinated manner. Meanwhile, the thumb exhibits a relatively higher degree of motion independence, reflecting its unique anatomical structure and functional role in grasping and manipulation tasks. These findings provide valuable insights for the design of dexterous robotic hands. Accordingly, the following section focuses on the simplification and implementation of the identified coupled joints to emulate natural hand functionality. The original kinematic data derived from this analysis are further employed to optimize the structure parameters of the proposed robotic hand, thereby enhancing both functional efficiency and mechanical feasibility.

\section{Mechanism Design}

The proposed dexterous hand takes the human hand as its design reference, aiming to replicate its anatomical structure, kinematic characteristics, and functional capabilities, while achieving comparable grasping and manipulation performance. As discussed above, the human hand exhibits strong inter-joint synergistic coordination, wherein multiple degrees of freedom follow coupled and predictable motion patterns. This synergistic characteristic is exploited to reduce the dimensionality of the actuation space while preserving the essential manipulation capabilities. Based on this reduced configuration, corresponding linkage and transmission mechanisms are proposed, resulting in a mechanical architecture that is both compact and straightforward to implement.

\begin{figure}[!t]
    \centering
    \includegraphics[width=1\linewidth]{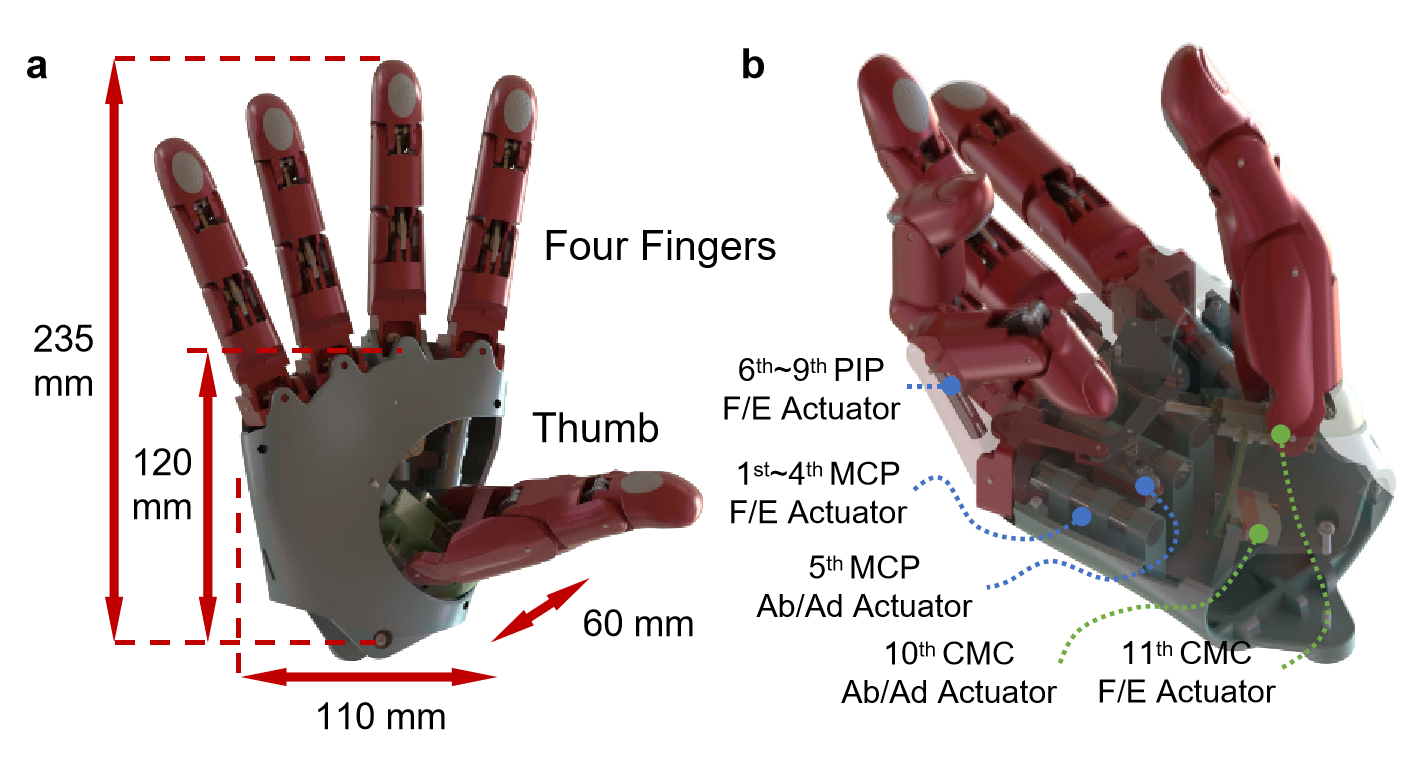}
    \caption{Overall structure of the SyLink Hand. (a) Three-dimensional model with key dimensions. (b) Actuator layout and joint distribution.}
    \label{FigureLabel3}
\end{figure}

\subsection{Hand Mechanism}

The hand comprises a palm and five fingers, with 19 joints actuated by 11 independent motors. The overall 3D model and spatial layout are illustrated in Fig.~\ref{FigureLabel3}a. The hand measures approximately 235mm in length, 110mm in width, and 60mm in depth, with a total mass of 520g. The fingertips can exert a peak force of 5N, and the hand supports a payload capacity exceeding 25N. By incorporating miniature motors, servo actuators, and 3D-printed structural components, the design achieves a total manufacturing cost around USD 400. Table I presents a quantitative comparison between the proposed hand and typical dexterous hands. The detailed component composition and associated costs are summarized in Table II.

To enable precise motion control and the generation of complex coupled trajectories, linkage mechanisms are extensively employed throughout the design \cite{Birglen2009TypeSO}, ensuring multi-joint coordination while maintaining structural compactness. The actuation scheme and joint distribution are shown in Fig.~\ref{FigureLabel3}b. Specifically, the thumb is actuated by a motor driving its coupled Flex/Ext motion and a servo governing its Abd/Add movement. For the remaining fingers, Flex/Ext is driven by two independent motors, while a single servo coordinates the lateral deviation of all three fingers simultaneously. Pressure and angular sensors are embedded within the fingers to provide real-time force and position feedback. An Arduino microcontroller is employed for sensor acquisition, actuator control, and communication, enabling closed-loop control of both position and force. The actuation, transmission, sensing, and control subsystems are fully integrated within the hand, resulting in a compact, efficient, and structurally flexible design.

\begin{table*}[t]
\centering
\caption{Quantitative comparison of representative dexterous robotic hands.}
\label{tab:comparison}

\renewcommand{\arraystretch}{1.25}
\setlength{\tabcolsep}{3pt}
\footnotesize

\resizebox{\textwidth}{!}{%
\begin{tabular}{p{2.5cm} c p{2.2cm} c p{1.8cm} c c p{3.8cm}}
\hline
\textbf{Platform}\rule{0pt}{5.0ex} &
\textbf{\shortstack{Actuated\\DOFs}} &
\textbf{Size (mm)} &
\textbf{Weight} &
\textbf{\shortstack{Approx. Cost}} &
\textbf{\shortstack{Fingertip\\Force}} &
\textbf{\shortstack{Payload}} &
\textbf{\shortstack{Transmission}} \\
\hline

DLR/HIT Hand II \cite{liu2008multisensory}
& 15
& Larger than human
& $\sim$1.5\,kg
& Very high
& 10\,N
& N/A
& Tendon + differential bevel gears \\

Shadow Hand \cite{sharma2014shadow}
& 20
& 448\,mm length
& $\sim$4.3\,kg
& $>$USD 50k
& $\sim$10\,N
& $\leq$5\,kg active
& Tendon-driven \\

ILDA Hand \cite{kim2021integrated}
& 15
& $218 \times 90 \times 56$
& 1.1\,kg
& High
& 34\,N
& 18\,kg
& Linkage-driven \\

RBO Hand 3 \cite{puhlmann2022rbo}
& 16
& Human scale
& $\sim$1.6\,kg
& High
& $\sim$8.3\,N
& $\sim$3.9\,kg passive
& Pneumatic soft actuation \\

Leap Hand \cite{shaw2023leap}
& 16
& $\sim 300 \times 100$
& $\sim$350\,g
& $\sim$USD 2k
& $\sim$5\,N
& N/A
& Direct geared servos \\

DexHand 021 \cite{yuan2025development}
& 12
& $296 \times 113 \times 56$
& $\sim$1\,kg
& $\sim$USD 16k
& $\sim$12\,N
& 5\,kg active
& Tendon-driven rope-elastic \\

% ORCA Hand \cite{christoph2025orca}
% & 17
% & Human scale
% & $\sim$1.3\,kg
% & $<$CHF 2k
% & N/A
% & $\sim$10.5\,kg passive
% & Tendon-driven \\

RUKA Hand \cite{zorin2025ruka}
& 11
& $\sim$180\,mm length
& $\sim$500\,g
& $\sim$USD 1.3k
& 2.74\,N
& $\sim$6\,kg passive
& Tendon, underactuated \\

Krysalis Hand \cite{basheer2025krysalis}
& 18
& $240 \times 92$
& 790\,g
& High
& 10\,N
& $>$4.5\,kg passive
& Integrated geared actuation \\

% CYJ Hand \cite{chai2025customize}
% & 22
% & Forearm-integrated
% & $\sim$750\,g
% & $<$USD 200
% & Low
% & $\sim$10\,kg passive
% & Tendon-driven \\

% Inspire RH56F1
% & 6
% & Human scale
% & 620\,g
% & $\sim$USD 5.5k
% & 15\,N
% & 30\,kg passive
% & Linear servo actuators \\

CasiaHand \cite{yan2025casiahand}
& 7
& $245 \times 93 \times 56$
& $\sim$890\,g
& Low
& 16.43\,N
& N/A
& Tendon, underactuated \\

ISyHand \cite{richardson2025isyhand}
& 18
& $255 \times 130 \times 38$
& $\sim$620\,g
& $\sim$USD 1.3k
& $\sim$5.5\,N
& $\sim$9\,kg passive
& On-joint servo-driven \\

% SX-hand \cite{chen2021design}
% & 12
% & Human-scale
% & 1.92\,kg
% & N/A
% & $>$15\,N
% & $\sim$20\,kg passive
% & Tendon, underactuated \\

X-hand \cite{xiong2016design}
& 4
& Human-scale
& N/A
& N/A
& $>$5\,N
& N/A
& Linkage, tendon, underactuated \\

\hline
\textbf{This Work}
& \textbf{11}
& $\mathbf{235 \times 110 \times 60}$
& \textbf{520\,g}
& \textbf{$\sim$USD 400}
& \textbf{$>$5\,N}
& \textbf{$>$2.5\,kg passive}
& \textbf{Linkage-driven + worm gear} \\
\hline

\end{tabular}%
}
\end{table*}

\begin{table}[htbp]
\centering
\caption{Manufacturing Cost Per Hand}
\label{tab:Cost}
\begin{tabular}{lccc}
\toprule
\textbf{Item} & \textbf{Unit Cost} & \textbf{Number} &\textbf{Total}\\ 
\midrule
MCP F/E motor  & 40  & 4 & 160\\
CMC F/E motor  & 5  & 1 & 5\\
PIP F/E motor   & 15  &  4 & 60\\
Lateral motor  & 35  & 2 & 70\\
Worm gear & 1.5  & 5 & 7.5\\
FSR film sensor & 3.5  & 5 & 17.5\\
Angle sensor & 1.2  & 5 & 6\\
3D-printed parts & 45  & - & 45\\ 
PCB boards & 5  & 1 & 5\\
Arduino controller & 12  & 1 & 12\\
Dowel pins and screws & 6  & - & 6\\ 
Electronic wire & 6  & - & 6\\
% Worm gear & 1.5  & 7 & 10.5\\
% Bearing & 0.5  & 5 & 2.5\\
% 3D-printed parts & 50  &  \textbackslash & 50\\
% CNC machined parts & 80 & \textbackslash & 80 \\
% Dowel pins and screws & 6 &  \textbackslash & 6\\
% Hall effect encoder & 5  & 2 & 10\\
% PCB boards & 5  & 2 & 10\\
% ESP32 controller & 3  & 2 & 6\\
% Electronic wire & 1  & 10 & 10\\
\midrule
Total Cost &  &  & 400\\
\bottomrule
\end{tabular}
\end{table}

\subsection{Finger Mechanism}

\begin{figure*}[!htbp]
    \centering
    \includegraphics[width=1\linewidth]{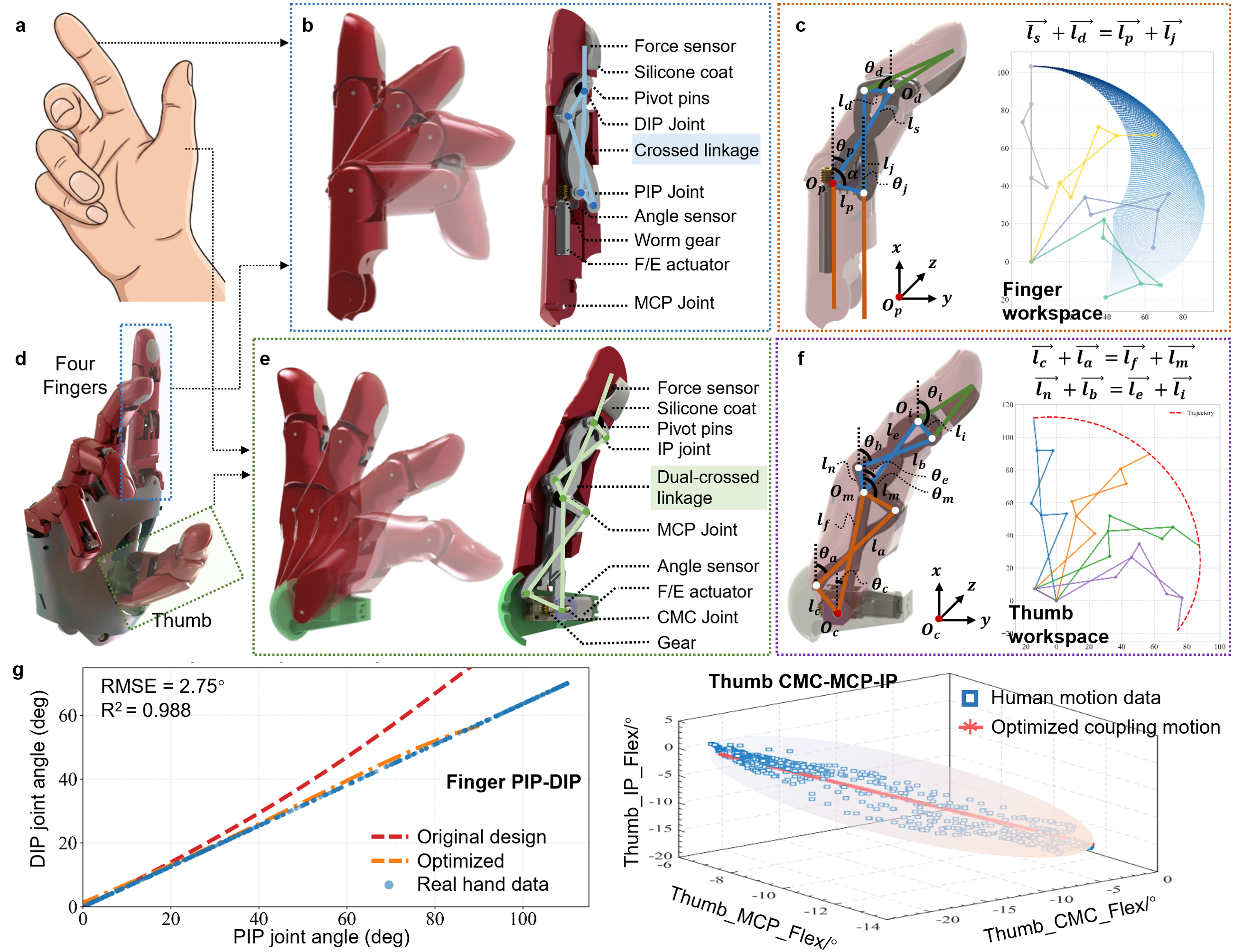}
    \caption{Finger and thumb mechanisms. (a) Human hand as morphological reference. (b) Finger configuration and module components. (c) Kinematic schematic and workspace trajectory of the finger mechanism. (d) Assembled hand prototype. (e) Thumb configuration and module components. (f) Kinematic schematic and workspace trajectory of the thumb mechanism. (g) Linkage optimization results against motion capture reference data.}
    \label{FigureLabel4}
\end{figure*}

Given the anatomical and kinematic similarity of the four fingers (excluding the thumb), they are designed as identical modular structures, which simplifies fabrication and reduces cost. These fingers are connected to the palm via base joints and mounted at varying heights to replicate the natural postural alignment of the human hand. As established above, the PIP and DIP joints exhibit a strong linear kinematic correlation, making them well-suited for coupled actuation. A crossed four-bar mechanism \cite{yoon2017underactuated} is therefore employed to link these two joint segments, enabling coordinated and mechanically consistent motion between them. To prevent undesired kinematic coupling between the MCP and PIP joints, the actuator is positioned inside the proximal phalanx, which also makes effective use of the available internal space within the finger. The motor output is transmitted to the connecting rod via a worm gear system, providing a compact transmission structure with inherent position self-locking capability. The structure and motion of the finger are illustrated in Fig.~\ref{FigureLabel4}b.

The thumb, as the most functionally critical digit of the human hand, enables stable and versatile grasping through its capacity to oppose all four fingers across a wide range of postures \cite{wang2017thumb}. Due to its unique anatomical structure and spatial orientation, the thumb necessitates a mechanical design distinct from the other fingers. In the human hand, the first metacarpal bone provides the thumb with both rotational and flexion capabilities, allowing it to perform complex motions necessary for dexterous manipulation and diverse grasping tasks. Hence, to preserve the reconfigurable relationship between the thumb and the palm, a three-joint biomimetic structure is adopted. Given the strong kinematic correlation between the Flex/Ext movements across the thumb joints, underactuated coupled actuation is both biomechanically justified and kinematically sufficient to reproduce the full range of opposition motion. Accordingly, thumb flexion is driven by a single motor located at the CMC base joint, whose output is transmitted through two stacked crossed four-bar linkages to produce coordinated, proportional motion across all three joints. Figure~\ref{FigureLabel4}e illustrates the coupled thumb Flex/Ext motion and the embedded linkage mechanism. Furthermore, lateral movement is independently actuated by a servo motor embedded within the palm, which drives a planar four-bar mechanism to adjust the angular orientation of the thumb relative to the palm, as illustrated in Fig.~\ref{FigureLabel5}d. The combination of these two degrees of freedom allows the thumb to generate a large motion workspace, enabling effective opposition and contact with each of the other fingers for precision grasping. 

The schematic diagrams of the finger and thumb mechanisms are presented in Fig.~\ref{FigureLabel4}c and~\ref{FigureLabel4}f respectively, both of which employ crossed four-bar linkages to realize underactuated coupled joint motion. Therefore, to quantitatively assess their kinematic performance, the complex linkage assembly is simplified into an equivalent four-bar linkage model, with the four-finger mechanism adopted as the representative case for analysis. As shown in Fig.~\ref{FigureLabel4}c, the mechanism satisfies the closed-loop vector equations:
\begin{equation}
\vec{l_s} + \vec{l_d} = \vec{l_p} + \vec{l_j}
\end{equation}

In the established coordinate frame, this relationship can be expressed as:
\begin{equation}
\begin{cases}
l_p \cos\alpha + l_j \cos\theta_j= l_s \cos\theta_p + l_d \cos\theta_d, \\
l_p \sin\alpha + l_j \sin\theta_j= l_s \sin\theta_p - l_d \sin\theta_d,
\end{cases}
\end{equation}

where $l_{p}$, $l_{j}$, $l_{s}$, $l_{d}$ denote the lengths of the respective links, $\theta_{p}$ and $\theta_{d}$ correspond to the rotation angles of the PIP and DIP joints respectively, $\alpha$ denotes the angular position of the fixed link, and $\theta_{j}$ represents the rotation angle of link $l_{j}$. With the known value of $l_{p}$, $l_{j}$, $l_{s}$, $l_{d}$, and $\alpha$, the kinematic relationship between the PIP and DIP joints can be quantitatively calculated.
\begin{equation}
\begin{split}
2l_s l_d \cos(\theta_p + \theta_d) - 2l_p l_d \cos(\theta_d + \alpha) \\ 
= l_j^2 - l_s^2 - l_d^2 - l_p^2 + 2l_p l_s \cos(\theta_p - \alpha)
\end{split}
\end{equation}

Eq. (6) can be simplified to:
\begin{equation}
A\cos\theta_d + B\sin\theta_d = C
\end{equation}

where:
\begin{align}
A &= 2l_s l_d \cos\theta_p - 2l_p l_d \cos\alpha \\
B &= 2l_s l_d \sin\theta_p + 2l_p l_d \sin\alpha \\
C &= l_j^2 - l_s^2 - l_d^2 - l_p^2 + 2l_p l_s \cos(\theta_p - \alpha)
\end{align}

The analytical solution is therefore given by:
\begin{equation}
\theta_d = \text{arctan2}(B,A) \pm \arccos\left(\frac{C}{\sqrt{A^2+B^2}}\right)
\end{equation}

The two solutions correspond to the two possible assembly configurations of the mechanism, and the physically valid solution is selected based on the kinematic constraints. 
With the analytical relationship between $\theta_p$ and $\theta_d$ established, the geometric parameters of the linkage mechanism, namely $l_p$, $l_j$, $l_s$, $l_d$, and $\alpha$, are optimized to best reproduce the natural coupled motion patterns observed in human finger kinematics. The optimization is formulated as a nonlinear least-squares problem, where the objective is to minimize the discrepancy between the analytically computed $\theta_d$ and the reference joint angle trajectories extracted from the motion capture dataset. The objective function is defined as:
\begin{equation}
\min_{\mathbf{x}} f(\mathbf{x}) = \sum_{i=1}^{N} \left( \theta_d^{\text{computed}}(\theta_p^i, \mathbf{x}) - \theta_d^{\text{ref},i} \right)^2
\end{equation}

where $\mathbf{x} = [l_p, l_j, l_s, l_d, \alpha]^\top$ denotes the vector of 
design parameters to be optimized, $N$ is the number of data points, 
$\theta_p^i$ and $\theta_d^{\text{ref},i}$ are the $i$-th reference PIP and 
DIP joint angles from the motion capture dataset, and 
$\theta_d^{\text{computed}}(\theta_p^i, \mathbf{x})$ is the analytically 
computed DIP angle given $\theta_p^i$ and the current parameter vector 
$\mathbf{x}$.

To ensure mechanical feasibility and anatomical compatibility, the following constraints are imposed on the design parameters, where the bounds are determined by anthropometric measurements of human finger dimensions:
\begin{equation}
\mathbf{x}_{\min} \leq \mathbf{x} \leq \mathbf{x}_{\max}
\end{equation}

Additionally, Grashof's condition is enforced to guarantee continuous relative motion of the linkage:
\begin{equation}
l_{\min} + l_{\max} \leq l_{\text{others},1} + l_{\text{others},2}
\end{equation}

Particle Swarm Optimization (PSO) is employed to solve this constrained 
nonlinear optimization problem, owing to its demonstrated effectiveness in 
handling high-dimensional, non-convex design spaces. In PSO, a swarm of $M$ 
particles explore the parameter space, where each particle $k$ is characterized 
by its position $\mathbf{x}_k$ and velocity $\mathbf{v}_k$. The update rules 
are given by:
\begin{align}
\mathbf{v}_k^{t+1} &= w\mathbf{v}_k^t + c_1 r_1 (\mathbf{p}_k^t - \mathbf{x}_k^t) + c_2 r_2 (\mathbf{g}^t - \mathbf{x}_k^t) \\
\mathbf{x}_k^{t+1} &= \mathbf{x}_k^t + \mathbf{v}_k^{t+1}
\end{align}

where $w$ is the inertia weight, $c_1$ and $c_2$ are cognitive and social 
acceleration coefficients, $r_1$ and $r_2$ are random numbers uniformly 
distributed in $[0,1]$, $\mathbf{p}_k^t$ is the personal best position of 
particle $k$, and $\mathbf{g}^t$ is the global best position of the swarm. 
The optimization terminates upon convergence or upon reaching the maximum 
number of iterations.

The optimization results are illustrated in Fig.~\ref{FigureLabel4}g, where the fitted trajectories are compared against the reference motion capture data. The optimization framework is further applied to the thumb coordinated linkage mechanism, enabling the corresponding linkage parameters to be systematically identified such that the resulting trajectories closely replicate the natural synergistic motion patterns of the human hand.

\subsection{Coordinated Lateral Motion Mechanism}

Although human fingers are capable of performing independent Abd/Add movements, such motion is rarely isolated in activities of daily living (ADL). Instead, the fingers typically abduct and adduct simultaneously as a coordinated group \cite{xu2013design, abayasiri2020under, basheer2025krysalis}. As aforementioned, the index, middle, ring, and little fingers tend to exhibit synchronized motion in natural movements. Moreover, given that the middle finger exhibits comparatively little lateral displacement relative to the others, it is rigidly fixed to the palm in this design. Based on this biomechanical observation and the objective of reducing structural complexity, a single-DOF planar linkage mechanism is adopted to drive the coordinated lateral motion of the index, ring, and little fingers, thereby reproducing natural hand postures.

\begin{figure*}[!htbp]
    \centering
    \includegraphics[width=1\linewidth]{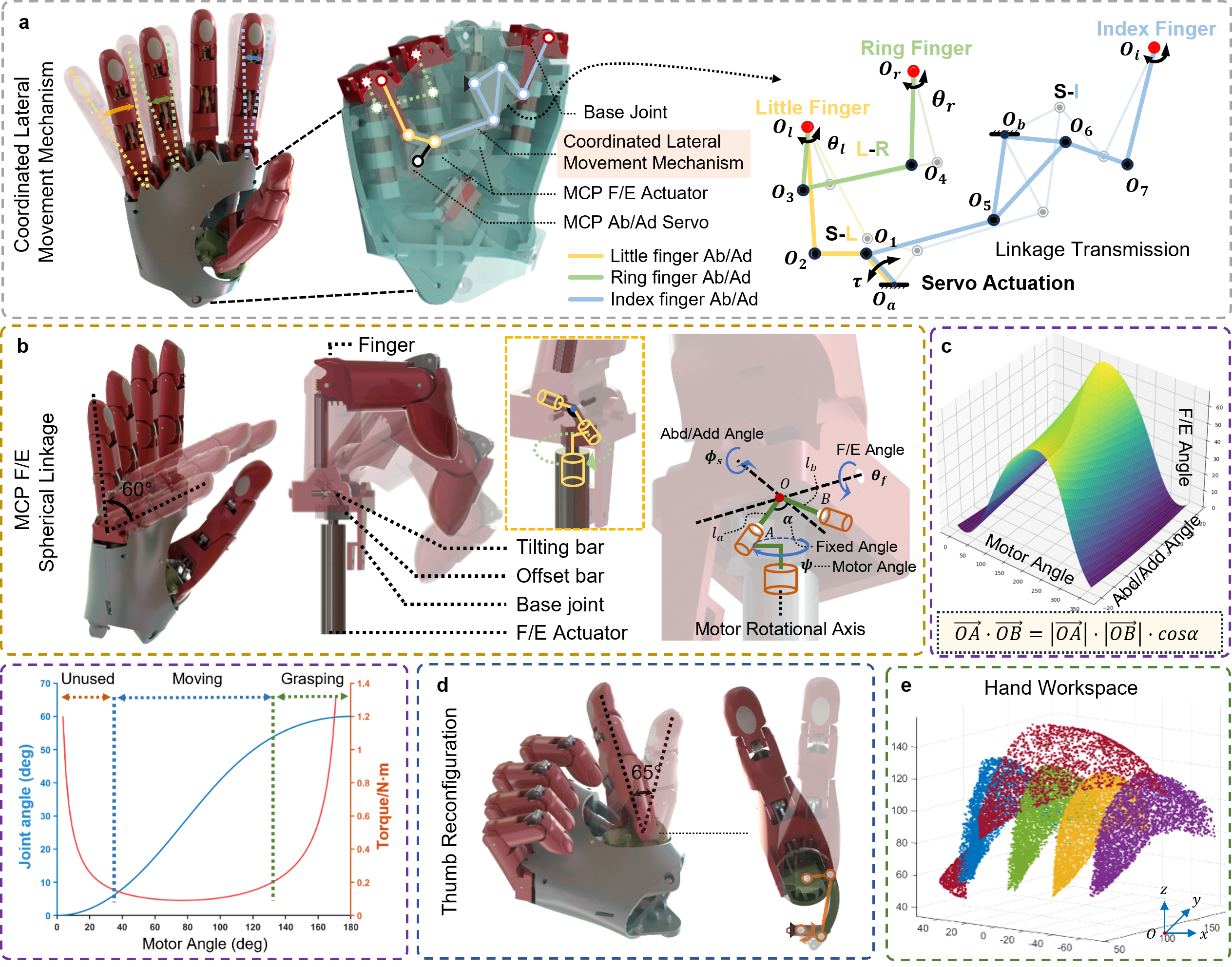}
    \caption{Transmission mechanisms and hand workspace. (a) Single-actuator coordinated lateral motion mechanism. (b) Spherical four-bar linkage mechanism for decoupled 2-DOF actuation at the MCP joint. (c) Computed relationship between motor angle and joint angle. (d) Planar four-bar linkage for independent thumb Abd/Add actuation. (e) Three-dimensional reachable workspace of the hand.}
    \label{FigureLabel5}
\end{figure*}

The mechanism for Abd/Add motion is illustrated in Fig.~\ref{FigureLabel5}a. A micro servo motor (KST/X06), embedded within the palm, is employed to actuate the lateral motion of the fingers. For the ring and little fingers, the motor output is transmitted to the base joint of the little finger through a planar four-bar linkage. The base joints of the ring and little fingers are further interconnected, forming an additional equivalent four-bar linkage. Through this coupled configuration, the motor drives both fingers to rotate laterally in the same direction. On the index finger side, the base joint is connected to the motor via a triangular linkage and a pair of rods, forming a five-bar linkage composed of two stacked four-bar linkages. This configuration reverses the direction of actuation, enabling the index finger to move laterally in the opposite direction to the ring and little fingers. By optimizing the geometric parameters of the linkages, the proposed mechanism effectively reproduces the natural Abd/Add kinematics of the human hand, allowing simultaneous lateral motion of the three fingers using only a single actuator. Under maximum abduction, the little finger achieves the largest lateral angle of approximately $25^\circ$, whereas the index and ring fingers exhibit comparable lateral angles of approximately $15^\circ$.

% Notably, the lateral motion of the index finger plays a critical role in stabilizing and supporting grasped objects. Here, to enhance its stabilization capability, the linkage mechanism is specifically designed with a large positive transmission angle ($\delta$), ensuring efficient torque transmission from the actuator to the finger. Conversely, the reverse transmission angle ($\eta$) is minimized to resist the external force that could displace the finger position. This configuration effectively locks the finger in its intended position, providing increased resistance against unintended movements. By leveraging the mechanical advantage, this approach enhances the stability and reliability of the finger during grasping and manipulation tasks.

\subsection{Spatial Spherical Four-bar Mechanism}

The incorporation of Abd/Add movements alters the spatial relationship between the MCP base joint and the motor output shaft responsible for Flex/Ext. Consequently, conventional transmission mechanisms, such as gears or worm gears, are unable to maintain effective meshing transmission across varying lateral positions. To address this, the DLR/HIT Hand II \cite{liu2008multisensory} employs a differential bevel gear mechanism, in which two motors operate synchronously or differentially to achieve decoupled Flex/Ext and Abd/Add movements. In contrast, the ILDA Hand \cite{kim2021integrated} adopts a linkage-driven mechanism that integrates parallel and serial kinematic chains to realize full 2-DOF motion at the MCP joint. Specifically, two prismatic–spherical–spherical (PSS) chains driven by two palm-mounted linear actuators independently control Flex/Ext and Abd/Add through differential and common-mode displacements. Nevertheless, both schemes introduce considerable mechanical complexity and a large spatial footprint. More importantly, they necessitate the simultaneous coordinated control of two motors to drive the MCP joint, which inherently precludes the realization of underactuated Abd/Add synergy discussed above. Therefore, a specialized mechanism that simultaneously accommodates and effectively decouples these two degrees of freedom within a compact form is required.

Here, we propose a spatial spherical four-bar linkage mechanism aimed at achieving effective decoupled 2-DOF motion within a compact spatial configuration. As shown in Fig.~\ref{FigureLabel5}b, this mechanism mainly consists of an offset bar and a tilting bar. Upon actuation, the motor's rotational motion is transmitted to the offset bar, which subsequently modulates the inclination angle of the tilting link. The finger joint, mounted on the upper surface of the tilting link, undergoes controlled Flex/Ext movement accordingly. The two rotational axes of the mechanism are mutually orthogonal and intersect at the center of the tilting link, denoted as point $O$, thereby constraining all key points of the mechanism to move on spherical surfaces centered at that point. This inherent geometric characteristic enables fully decoupled, independent, and mechanically simplified actuation of Flex/Ext and Abd/Add motions, which is otherwise difficult to achieve with conventional transmission mechanisms.

In the schematic diagram of the structure shown in Fig.~\ref{FigureLabel5}b, the two endpoints of the tilting link are defined as points $A$ and $B$, respectively. The position of point $A$ is related to the motor rotation angle $\psi$, whereas the position of point $B$ is determined by the Abd/Add angle $\phi_{s}$ and Flex/Ext angle $\theta_{f}$ of the finger. Meanwhile, the vectors $\mathbf{OA}$ and $\mathbf{OB}$ maintain a fixed spatial angle $\alpha$ between them. By design, when the motor angle is zero, the finger is oriented vertically. Thus, the coordinates of $\mathbf{OA}$ and $\mathbf{OB}$ can be expressed as:
\begin{equation}
\mathbf{OA} = \left( -l_a \cos\alpha \sin\psi,\; -l_a \cos\alpha \cos\psi,\; -l_a \sin\alpha \right),
\end{equation}
\begin{equation}
\mathbf{OB} = \left( l_b \sin\phi_s ,\; -l_b \cos\theta_f \cos\phi_s,\; l_b \sin\theta_f \cos\phi_s \right),
\end{equation}

where $l_{a}$ denotes the length of vector \(\boldsymbol{OA}\), $l_{b}$ denotes the length of vector \(\boldsymbol{OB}\). 

The finger joint Flex/Ext angle, Abd/Add angle, and motor rotation angle can be derived from the geometric constraints of the mechanism and are expressed by the following equation:
\begin{align}
\cos\alpha &= \frac{\boldsymbol{OA} \cdot \boldsymbol{OB}}{|\boldsymbol{OA}|\,|\boldsymbol{OB}|} = \cos\alpha \cos\psi \cos\theta_f \cos\phi_s \nonumber \\
&\qquad -\cos\alpha \sin\psi \sin\phi_s  - \sin\alpha \sin\theta_f \cos \phi_s,
\label{eq:angle_relation}
\end{align}

which leads to the following constraint equation:
\begin{equation}
\cos\phi_s \cos\psi \cos\theta_f
- \tan\alpha \cos\phi_s \sin\theta_f
= 1 + \sin\phi_s \sin\psi
\end{equation}

The motor rotates within the range $[-\pi,\pi]$. Therefore, for a given motor rotation angle $\psi$ and measured Abd/Add angle $\phi_{s}$, the joint flexion angle can be obtained via forward kinematics as:
\begin{equation}
\begin{split}
\theta_f = \pm \arccos\left(\frac{1+\sin\phi_s\sin\psi}{\sqrt{\cos^2\phi_s\cos^2\psi+\tan^2\alpha\cos^2\phi_s}}\right) \\ + \arctan\left(\frac{-\tan\alpha}{\cos\psi}\right),\quad \theta_f \in [0,\ \frac{\pi}{2}].
\end{split}
\end{equation}

For given target bending angle $\theta_{f}$ and lateral deflection angle $\phi_{s}$, the required motor angle $\psi$ can be derived as:
\begin{equation}
\begin{split}
\psi =  \pm \arccos\!\left(\frac{\cos\alpha + \sin\alpha \sin\theta_f \cos\phi_s}{\cos\alpha \sqrt{\cos^2\theta_f \cos^2\phi_s + \sin^2\phi_s}}\right) \\
       + \operatorname{arctan2}\!\left(-\sin\phi_s,\ \cos\theta_f \cos\phi_s\right),\quad \psi \in [-\pi,\ \pi].
\end{split}
\end{equation}

The relationship among these three angles is computed and visualized in Fig.~\ref{FigureLabel5}c. It can be observed that the flexion angle of the finger joint increases in proportion to the motor's rotational angle, reaching a maximum of approximately 60°. The Abd/Add angle has only a minor influence on the Flex/Ext angle, effectively achieving a substantial decoupling between the two degrees of freedom. In addition, the mechanism exhibits nonlinear characteristics in both motion transmission and torque generation. Specifically, the output torque decreases initially and subsequently increases throughout the motion cycle, enabling rapid finger closure during the approach phase while providing greater force output for stable grasping as contact with the object is established. Furthermore, the mechanism exhibits a self-locking characteristic, allowing it to maintain its position against external disturbances without continuous actuation. Collectively, these advantages enable the spherical four-bar mechanism to compactly achieve motion decoupling while providing both high dexterity and high load-bearing capability in anthropomorphic robotic fingers.

\begin{figure*}[!htbp]
    \centering
    \includegraphics[width=1\linewidth]{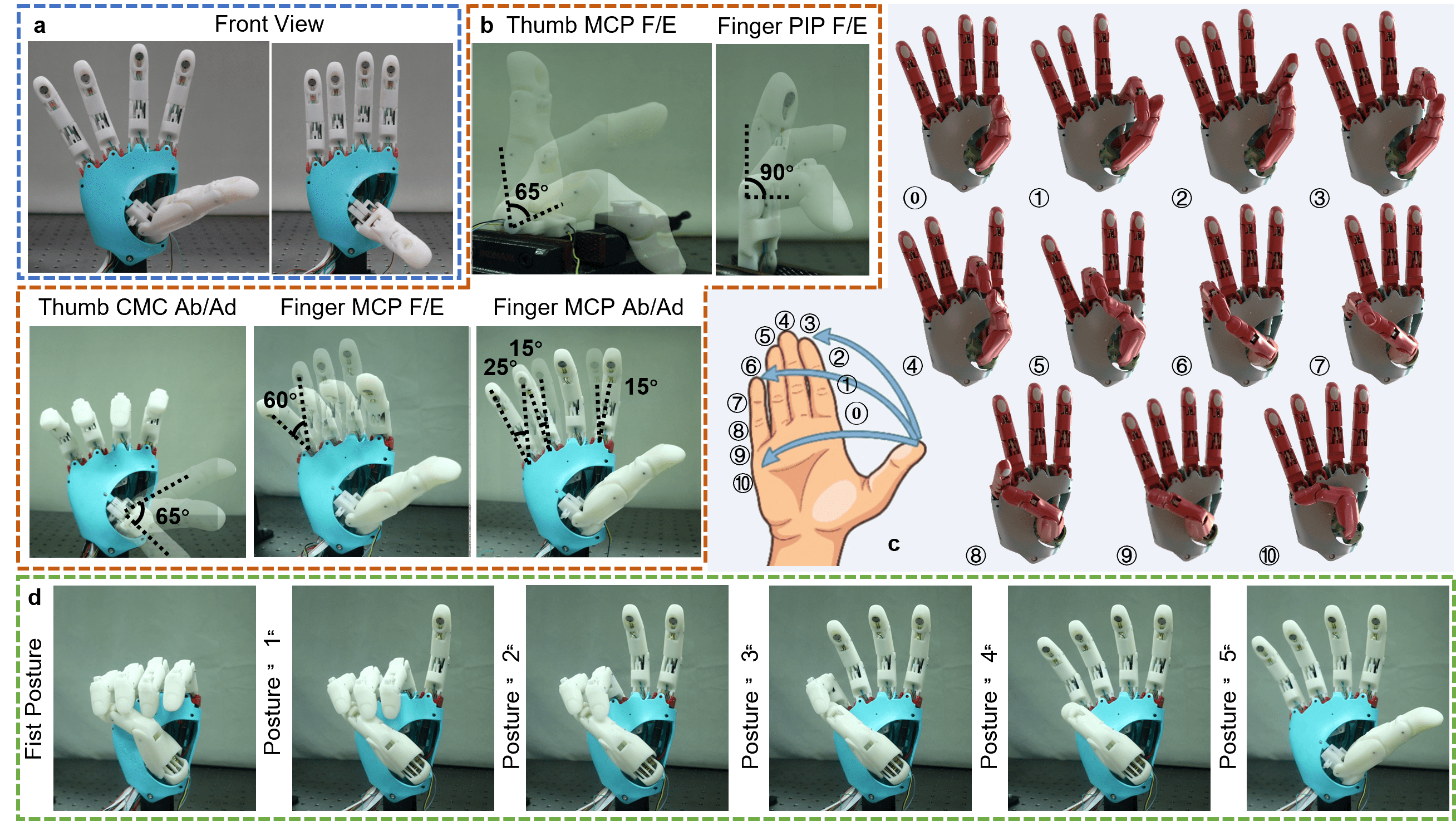}
    \caption{Motion dexterity evaluation. (a) Fabricated prototype. (b) Joint ranges of motion across all degrees of freedom. (c) The hand achieves the highest score in the Kapandji opposability test. (d) The hand gestures numbers 1–5 through coordinated multi-joint motion.}
    \label{FigureLabel6}
\end{figure*}

\subsection{Workspace}

Based on the defined joint ranges of motion, forward kinematic model, and spatial joint layout, the reachable workspace of each finger and the hand as a whole can be analytically derived and visualized. As illustrated in Fig.~\ref{FigureLabel5}e, both the four fingers and the thumb exhibit motion trajectories and workspace ranges comparable to those of the human hand. More importantly, the workspace of the thumb intersects with those of the other fingers, thus enabling effective opposable-thumb motions and versatile fingertip interactions. This overlap is critical for enabling stable pinch grasps and dexterous manipulation, as it ensures the thumb can establish reliable opposition against each finger across varying object sizes and grasp configurations.

\subsection{Integrated Sensors}

The perception system of our dexterous hand is designed to provide comprehensive and real-time feedback on the hand's motion and its interaction with objects. Specifically, rotation angle sensors are embedded at the finger joints to monitor the joint angles. For MCP joints actuated by motors with encoders, the angular information is derived by integrating the encoder data over time. This setup ensures accurate measurement of joint angles and closed-loop control over finger movements, contributing to the overall control and coordination of the hand's movements.

In addition to joint angle sensors, the dexterous hand is equipped with force sensors at the fingertip. These sensors are designed to measure the interaction forces between the hand and objects, providing essential feedback on the contact conditions. Specifically, force sensing resistor (FSR) film sensors are employed and embedded within the distal phalanx of each finger, with the external surface coated in silicone to ensure compliant contact and enhance frictional interaction. The fingertip force measurements are fed back into the control system in real time, enabling the hand to dynamically adjust its grip strength and finger posture to maintain stable and secure object interaction.

\section{Performance Evaluation}

To evaluate the performance of the proposed prototype, a series of experiments were conducted. The tests assessed the hand across four key aspects: motion dexterity, force and load-bearing characteristics, grasping and in-hand manipulation of a diverse range of objects, and tool operation for task-specific applications.

% \begin{table}[t]
% \centering
% \caption{Comparison of functional range of motion (RoM) between the proposed hand and the human hand\cite{barakat2013range, bain2015functional}.}
% \label{tab:rom_comparison}
% \renewcommand{\arraystretch}{1.2}
% \begin{tabular}{cccc}
% \hline
% \textbf{Part} & \textbf{Joint/Category} & \textbf{Our Hand ($^\circ$)} & \textbf{Human Hand ($^\circ$)} \\
% \hline

% \multirow{4}{*}{Thumb}
% & CMC Abd/Add           & 90 & 73.1 \\
% & CMC Flex/Ext          & 95 & 61.2 \\
% & MCP Flex/Ext          & 100  & 68.1 \\
% & IP Flex/Ext           & 80  & 100.0 \\
% & \textbf{Total RoM} & \textbf{365} & \textbf{302.4} \\
% \hline

% \multirow{5}{*}{Four fingers}
% & MCP Abd/Add           & 30 - 40  & 25 - 30 \\
% & MCP Flex/Ext          & 60 & 71.0 \\
% & PIP Flex/Ext          & 90  & 87.0 \\
% & DIP Flex/Ext          & 50  & 64.0 \\
% & \textbf{Total RoM} & \textbf{230 - 240} & \textbf{247 - 252} \\
% \hline

% \end{tabular}
% \end{table}

\subsection{Joint Range of Motion and Thumb Opposability}

The fabricated prototype of the dexterous hand is presented in Fig.~\ref{FigureLabel6}a, and the motions and corresponding joint ranges of all five fingers are illustrated in Fig.~\ref{FigureLabel6}b. Compared with the human hand, the proposed hand achieves comparable joint ranges across all fingers, enabling it to reproduce most major finger motions with high fidelity.

To evaluate the opposition performance of the thumb, we employed the Kapandji test \cite{KAPANDJI198667}, a clinically established method that assesses thumb dexterity by measuring its ability to reach a series of standardized positions across the hand. The results in Fig.~\ref{FigureLabel6}c show that, despite the underactuated design, the thumb achieved the highest Kapandji score, demonstrating that the simplified actuation preserved sufficient dexterity and opposition capability for diverse human-like grasping tasks. Furthermore, as illustrated in Fig.~\ref{FigureLabel6}d, the dexterous hand was commanded to perform gestures representing the number 1-5, thereby demonstrating the motion of all individual joints as well as the coordinated movement among multiple joints.

\subsection{Force and Load Performance}

\begin{figure}[!t]
    \centering
    \includegraphics[width=1\linewidth]{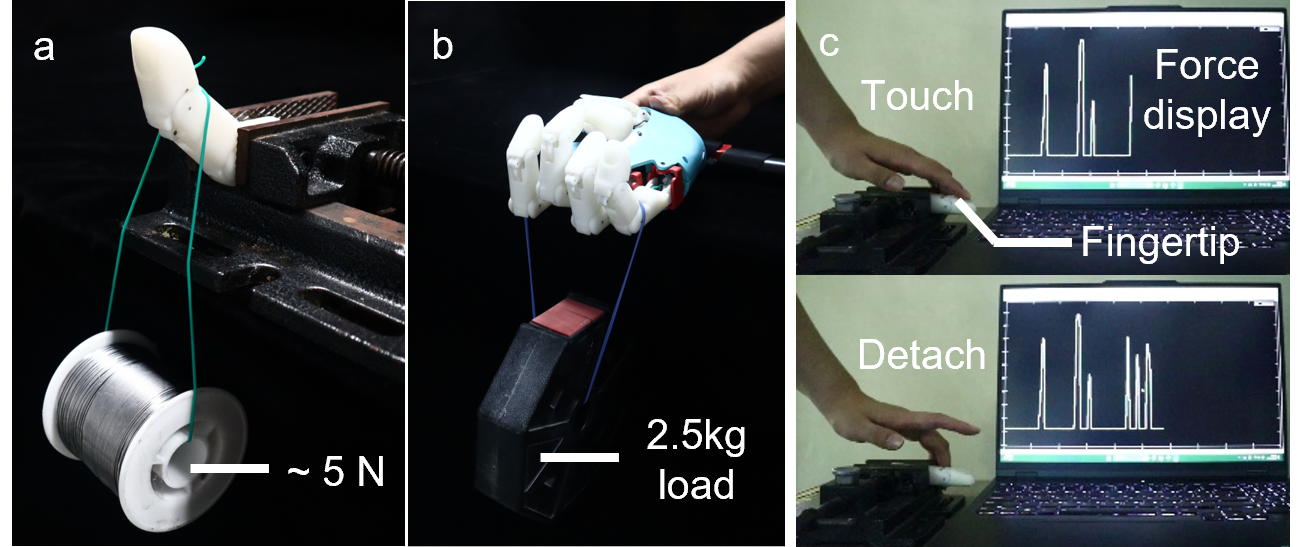}
    \caption{Force and load performance of the hand. (a) A single finger actively lifts a 500g object. (b) The hand passively holds a 2.5kg dumbbell. (c) The fingertip can sensitively perceive repeated contact forces.}
    \label{FigureLabel7}
\end{figure}

In this section, we characterize the force generation, load-bearing, and fingertip tactile sensing performance of the proposed anthropomorphic hand. First, to evaluate individual finger strength, a single finger was fixed in a bench vise and actuated to perform a full flexion motion. As shown in Fig.~\ref{FigureLabel7}a, the finger successfully lifted an object weighing approximately 500g, demonstrating that the miniature actuator and linkage transmission are capable of generating sufficient fingertip force for grasping tasks. In addition, all Flex/Ext movements of the finger joints are driven by
linkage or worm-gear mechanisms with inherent self-locking
characteristics, allowing the hand to passively maintain grasping states without continuous motor actuation. Benefiting from this characteristic, this fully 3D-printed hand was able to stably hold a 2.5kg dumbbell, as illustrated in Fig.~\ref{FigureLabel7}b, highlighting its load-bearing capability and structural robustness. Furthermore, each fingertip is integrated with a thin-film FSR tactile sensor capable of measuring contact forces in the range of 0-20N. Figure~\ref{FigureLabel7}c shows the sensor responses during repeated contact experiments, demonstrating consistent and sensitive fingertip force perception performance.

\begin{figure*}[!htbp]
    \centering
    \includegraphics[width=1\linewidth]{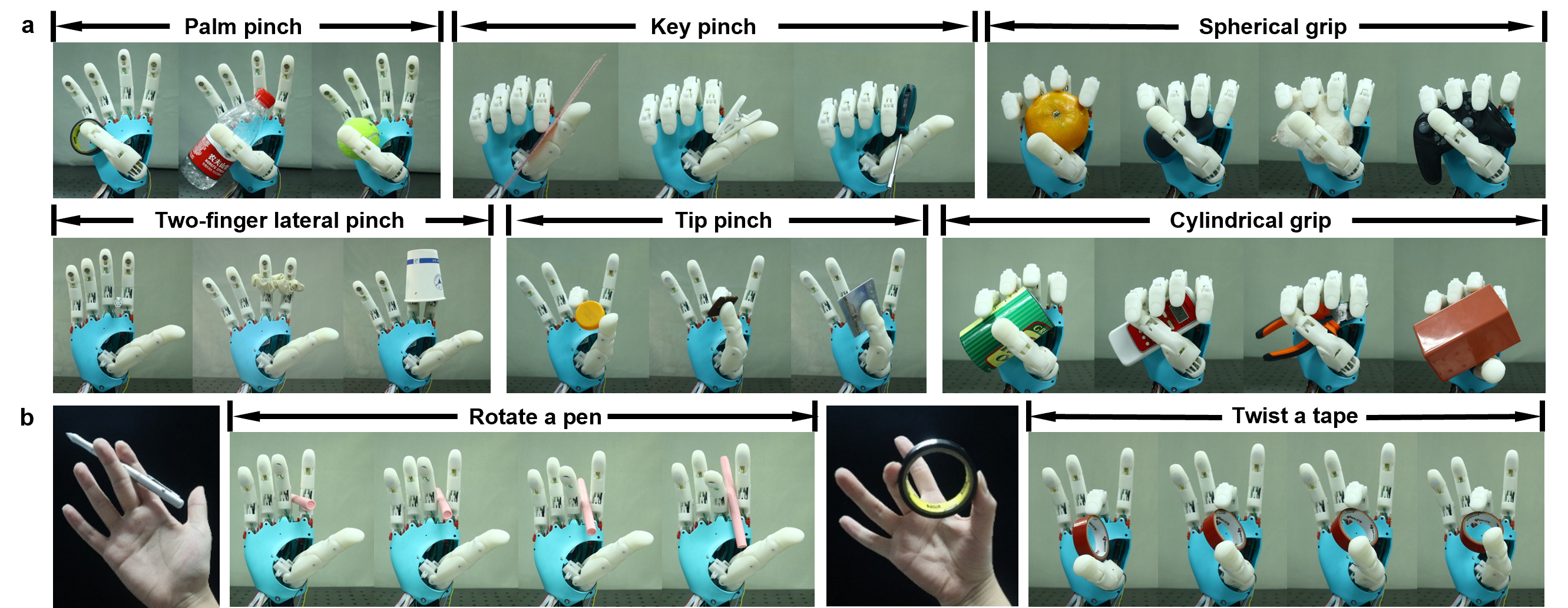}
    \caption{Grasping and in-hand manipulation. (a) Six representative grasp types executed on objects of varying shape, size, and texture. (b) In-hand pen repositioning and tape rotation through coordinated finger motion.}
    \label{FigureLabel8}
\end{figure*}

\begin{figure*}[!htbp]
    \centering
    \includegraphics[width=1\linewidth]{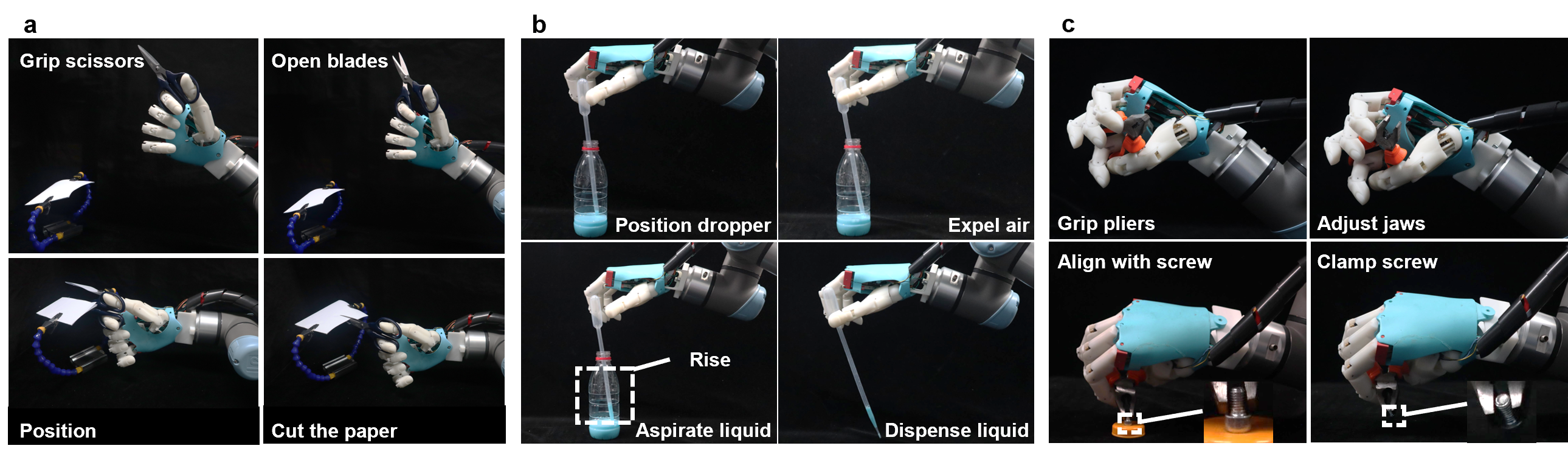}
    \caption{Tool operation demonstrations. (a) Paper cutting with scissors. (b) Liquid aspiration and dispensing with a pipette. (c) Screw gripping with pliers.}
    \label{FigureLabel9}
\end{figure*}

\subsection{Grasping and Manipulation Capacity Assessment}

The grasping capability of the robotic hand across objects of varying shape, weight, size, and surface texture is a key indicator of its overall functionality. To comprehensively evaluate this performance, a diverse collection of daily objects was selected as grasping targets, covering a wide range of real-world scenarios. Six representative grasping types were chosen for evaluation, including palm pinch, key pinch, spherical grip, two-finger lateral pinch, tip pinch, and cylindrical grip. The experimental results shown in Fig.~\ref{FigureLabel8}a demonstrate that the dexterous hand, with its anthropomorphic structure and high degrees of freedom design, successfully achieved stable grasps across all selected objects and effectively replicated the rich functional versatility of the human hand. These results validate the robustness and adaptability of the proposed design, confirming its potential for practical applications across a variety of grasping scenarios.

In-hand manipulation enables repositioning of objects within the hand, thereby facilitating more effective grasping, placing, and subsequent operations. To realize this functionality in this dexterous robotic hand, human hand motion data was recorded and used as a reference to drive coordinated finger movements on the dexterous hand. In the experiments, the hand grasped a pen using the middle and index fingers and adjusted its position through controlled flexion and lateral movements. Additionally, the hand successfully rotated a roll of tape while maintaining stable contact through a precise pinch. The results, as shown in Fig.~\ref{FigureLabel8}b, demonstrate that the hand prototype can effectively reposition objects to desired locations through the synchronized control of finger movements.

\subsection{Task-Oriented Tool Operation}

Achieving dexterous tool operation with the robotic hand requires real-time sensory feedback, adaptive force regulation, and precise sensorimotor coordination. To validate these capabilities in the proposed hand, three simple tool manipulation demonstrations were conducted. The prototype hand was mounted on a UR5e robotic arm, with motions executed by decomposing target movement sequences into discrete motion primitives. The three experiments were designed to specifically evaluate multi-finger coordination, precision control, and force-regulated manipulation.

In the paper-cutting experiment, the dexterous hand grasped the scissor handles with the thumb and index finger placed on opposing sides. The coordinated movement of these two digits allows the blades to open and close for cutting. Upon directing the scissors to the target position, the hand successfully severed the paper with a predetermined motion (Fig.~\ref{FigureLabel9}a). For liquid aspiration and transfer with a dropper, the hand demonstrated precise force modulation. As depicted in Fig.~\ref{FigureLabel9}b, a greater fingertip force was required to expel air during aspiration, whereas, during liquid transfer, the hand needed to maintain sufficient grip to prevent slippage while avoiding excessive pressure that could prematurely expel the contained liquid. In the pliers operation experiment, the hand securely gripped the pliers’ jaws with the thumb and index finger, while the ring and little fingers interlocked to hold the handle. As shown in Fig.~\ref{FigureLabel9}c, the opening and closing of the jaws were controlled by the extension and flexion of the fingers, thus effectively securing and lifting the screw from the desktop.

These experiments demonstrate the prototype hand's functionality in operating common tools and performing complex tasks. The results emphasize its multi-finger coordination, precise motion control, and fingertip force regulation capability, showcasing its potential for broad real-world applications.

\section{Conclusion}

In this work, we present a biomimetic, linkage-driven anthropomorphic hand that aims to replicate the dexterity, adaptability, and coordinated motion characteristics of the human hand. Inspired by the biomechanical principles of human hand synergies, the proposed design integrates 19 joints driven by 11 active actuators, achieving a balance between high dexterity and mechanical simplicity. To achieve efficient transmission of joint motions and the desired underactuated architecture, we combine spatial spherical linkage mechanisms, planar linkage transmissions, and worm-gear-based self-locking structures. Their geometric configurations are further optimized to achieve coordinated multi-joint motion, natural finger movement trajectories, and efficient force transmission, while maintaining a compact, lightweight, and highly integrated robotic system.

Experimental results demonstrate that the proposed hand achieves human-like motion dexterity and robust force generation capability. Furthermore, the prototype successfully performed versatile object grasping, dexterous in-hand manipulation, and a range of tool operations, validating its potential for practical deployment in real-world scenarios. More importantly, this work contributes a mechanically efficient and cost-effective design framework that bridges the gap between simple low-DOF grippers and highly expensive fully actuated dexterous hands by embedding biomechanical synergy directly into the kinematic architecture, achieving functional dexterity without proportional increases in mechanical complexity. This balance between capability and accessibility positions the proposed system as a practical platform for dexterity-demanding applications such as teleoperation and embodied robot learning.

Future work will focus on three directions. First, the sensing capabilities of the hand will be enhanced through the integration of distributed tactile sensing across the finger surfaces, enabling richer contact perception and more adaptive grasp control. Second, the mechanical structure will be further optimized and miniaturized to achieve a human-hand-like form factor while improving structural integration, portability, and actuation efficiency. Third, learning-based strategies will be explored to leverage the hand’s biomimetic kinematic structure, enabling the acquisition and generalization of dexterous manipulation skills across diverse objects and tasks.

%\begin{figure}[!t]
%    \centering
%    \includegraphics[width=1\linewidth]{fig/Correspondences.png}
%    \caption{Cooperativity among regions of the hand during object manipulation and grasp.}
%    \label{FigureLabel2}
%\end{figure}

%\begin{figure*}[!htbp]
%    \centering
%    \includegraphics[width=1\linewidth]{fig/Hand Finger Structure.png}
%    \caption{Demonstration of the hand structure. (A) Overall Structure of the Dexterous Hand. (B) Four-Fingered Structure. (C) Thumb Structure.}
%    \label{FigureLabel3}
%\end{figure*}

%\begin{figure}[!t]
%    \centering
%    \includegraphics[width=1\linewidth]{fig/Joint Structure_Vert.png}
%    \caption{Two-degree-of-freedom joint structure. (A) Coordinated lateral motion mechanism. (B) Spherical four bar linkage structure. (C) Structural diagram of the spherical linkage mechanism. (D) Nonlinear characteristic of the spherical four bar linkage for angle and torque.}
%    \label{FigureLabel4}
%\end{figure}

\bibliographystyle{IEEEtran}
\bibliography{reference.bib}

\begin{IEEEbiography}
[{\includegraphics[width=1in,height=1.25in]
{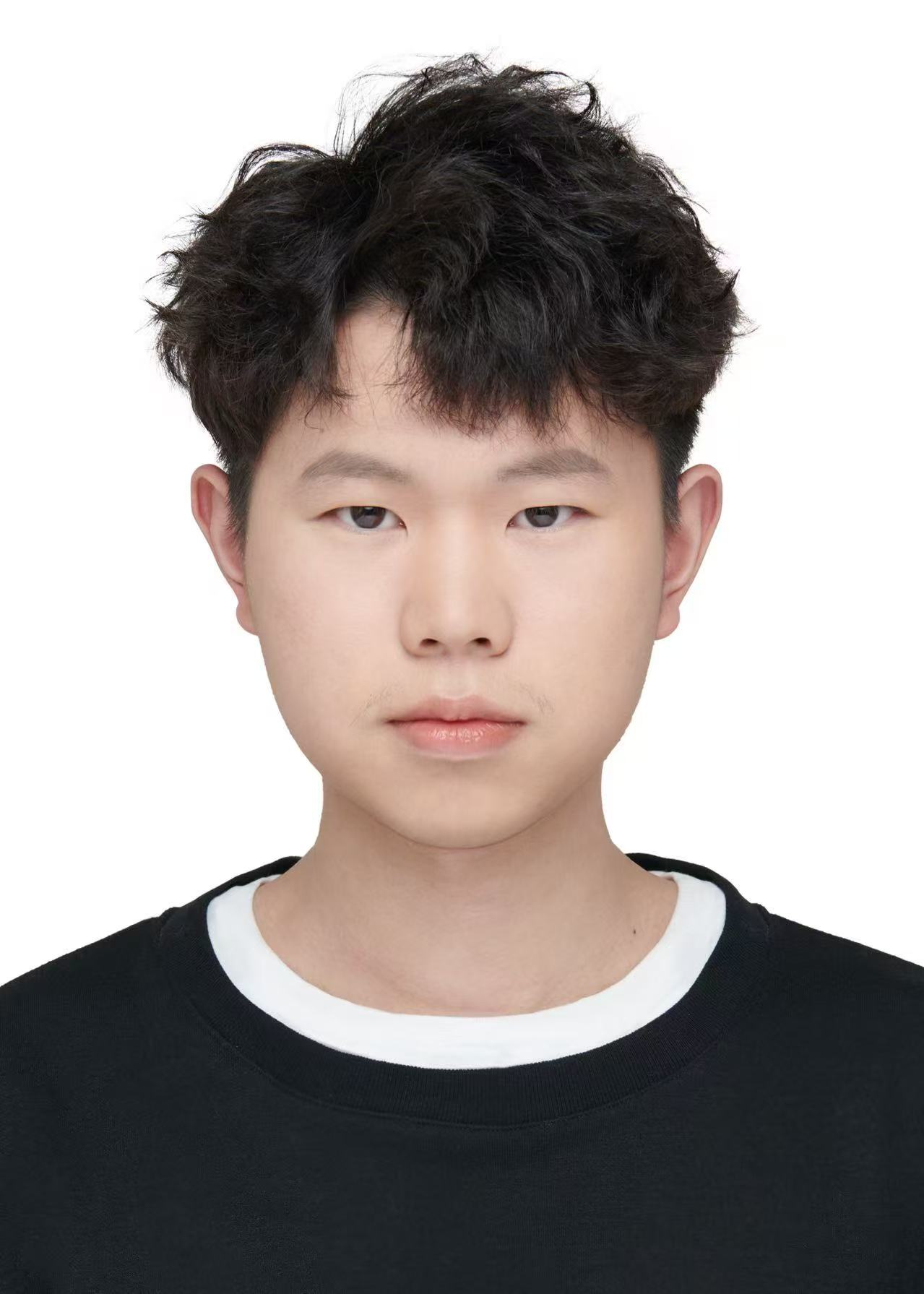}}]
{HaoWu} received the B.E. degree in Mechanical Engineering from Zhejiang University, China in 2024. He worked with the Grasp Lab, Zhejiang University, from 2023 to 2025. He is currently pursuing the Ph.D. degree with the Department of Mechanical Engineering, National University of Singapore, Singapore, under the supervision of Dr. Jianshu Zhou. His current research interests include mechanism design, soft robotics, and multimodal tactile perception. 
\end{IEEEbiography}

\begin{IEEEbiography}
[{\includegraphics[width=1in,height=1.25in]{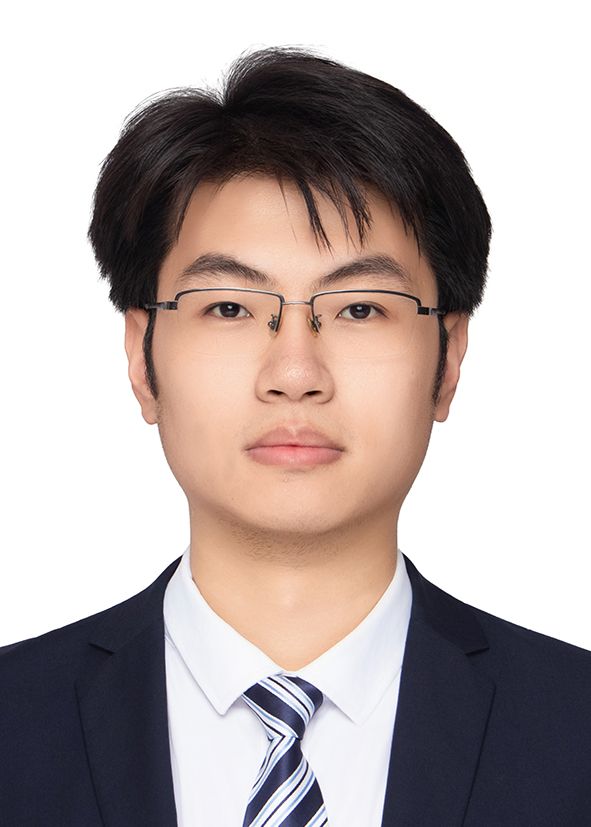}}]
{YanzheWang} received the B.S. degree in vehicle engineering from Jilin University, Changchun, China, in 2017 and obtained Ph.D. degree from the School of Mechanical Engineering, Zhejiang University, Hangzhou, China 2023. He is currently a post-doctoral fellow at Grasp Lab, Institute of Engineering, Aerospace Manufacturing School of Mechanical Engineering, Zhejiang University. His current research interests include robotic motion planning, multimodal perception and manipulation. 
\end{IEEEbiography}

\begin{IEEEbiography}[{\includegraphics[width=1in,height=1.25in]
{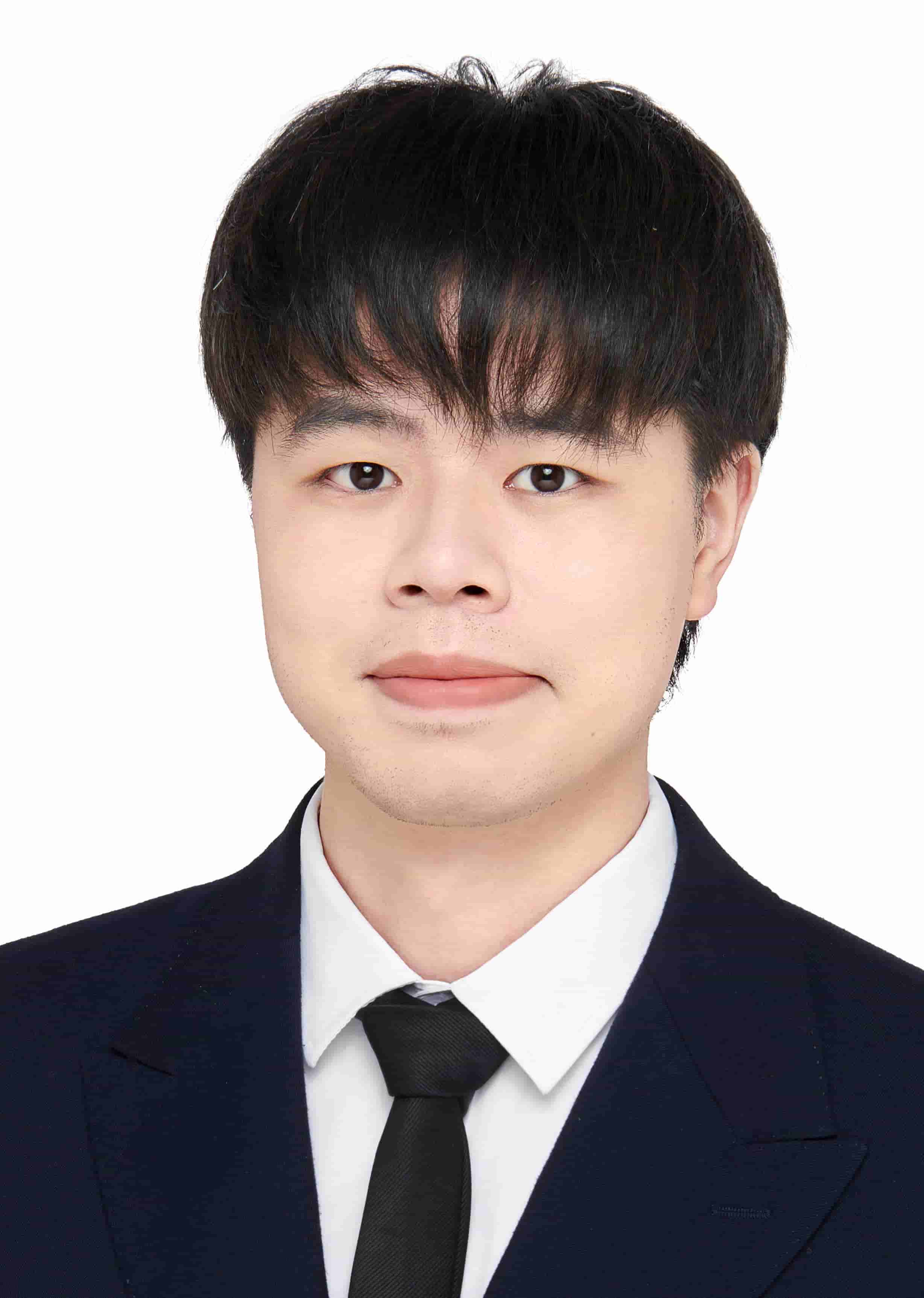}}]{YuFeng} received a B.Eng. degree in Measurement and Automation from Harbin Engineering University, Harbin, China, in 2020 and a Ph.D. degree in Mechanical Engineering from City University of Hong Kong, Hong Kong SAR, in 2024. He is currently working as a postdoctoral fellow at the City University of Hong Kong. His research interests include AI-enabled sensors, soft tactile sensors, and soft robotics.
\end{IEEEbiography}

\begin{IEEEbiography}[{\includegraphics[width=1in,height=1.25in]
{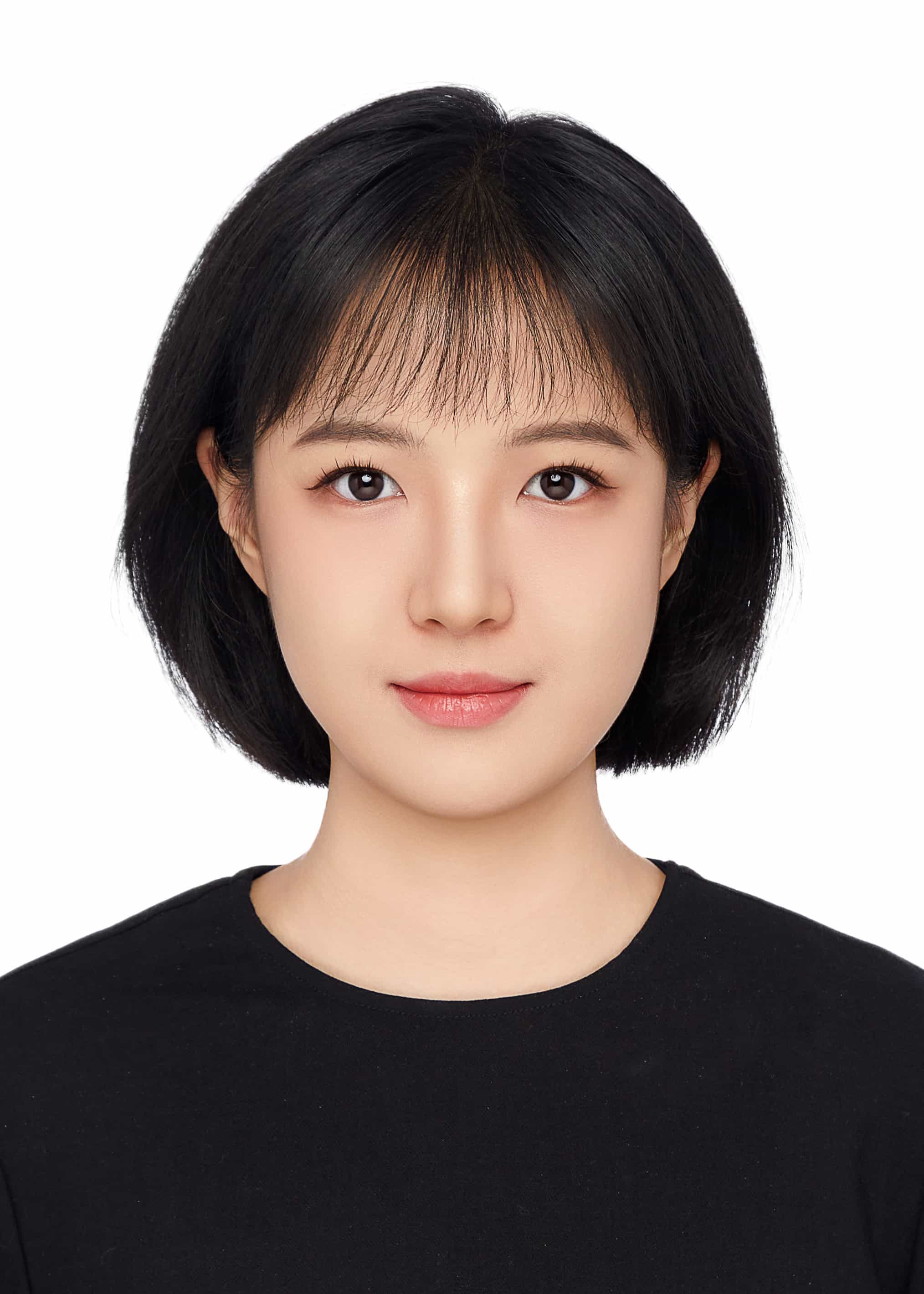}}]{Yitong Li} received the B.S. and M.S. degrees in mechanical engineering from the School of Power and Mechanical Engineering, Wuhan University, Wuhan, China, in 2022 and 2025, respectively. Her research interests include mechatronic systems, compliant mechanisms, robotics, and robotic mechanisms.
\end{IEEEbiography}

\begin{IEEEbiography}[{\includegraphics[width=1in,height=1.25in]
{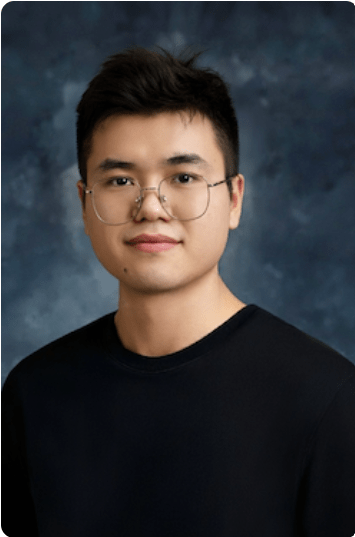}}]{Jingxiang Guo} is currently pursuing the Ph.D. degree at NUS, with the Language and Vision Lab (LV-Lab), under the supervision of Prof. Shuicheng Yan and co-supervision of Prof. Jianshu Zhou from the G\&M Lab. He received the B.Eng. degree in Automation from the Harbin Institute of Technology, Shenzhen (HITSZ), China, and the M.Comp. degree in Artificial Intelligence from the National University of Singapore (NUS), Singapore.
\end{IEEEbiography}

\begin{IEEEbiography}[{\includegraphics[width=1in,height=1.25in]
{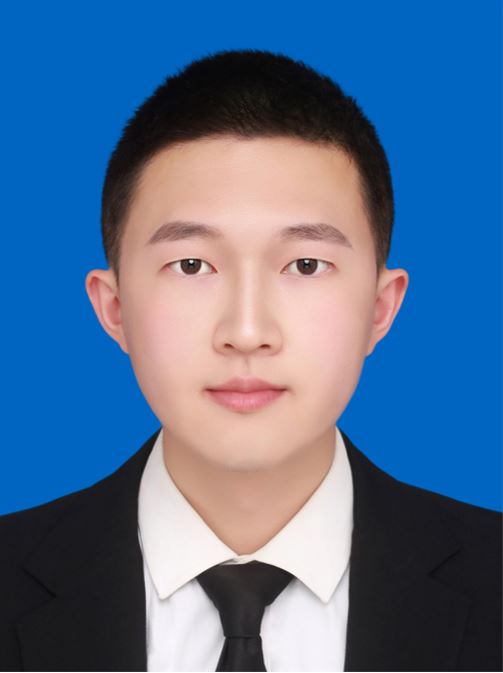}}]{Jian Liu} received his Ph.D. degree in 2025 at the National Engineering Research Center of Robot Visual Perception and Control Technology, Hunan University, Changsha, China. He is a Research Fellow at National University of Singapore. From 2023 to 2024, he was a Visiting PhD at the Department of Computer Science of the University of Western Australia. His current research interests include robotic manipulation, 3D machine vision, and object pose estimation. He served as a reviewer for more than 20 journals and conferences.
\end{IEEEbiography}

\begin{IEEEbiography}
[{\includegraphics[width=1in,height=1.25in]{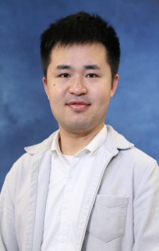}}]
{JianshuZhou} (Member, IEEE/ASME) received the Ph.D. degree in mechanical engineering from The University of Hong Kong in 2020. He is currently an Assistant Professor with the Department of Mechanical Engineering at the National University of Singapore. Prior to this, he was a Postdoctoral Scholar with the Mechanical Systems Control Laboratory, Department of Mechanical Engineering, University of California, Berkeley. He also served as a Research Assistant Professor at The Chinese University of Hong Kong. His research interests include robotics, mechatronics, robotic grasping and manipulation, novel actuation mechanisms, and dexterous robotic hands.
\end{IEEEbiography}

% \begin{IEEEbiography}
% [{\includegraphics[width=1in,height=1.25in]{bio/HuixuDong.png}}]
% {Huixu Dong} (S’17–M’18) received the B.Sc degree in mechatronics engineering from Harbin Institute of Technology in China, in 2013 and obtained  Ph.D. degree at Robotics Research Centre of Nanyang Technological University, Singapore 2018. He was a post-doctoral fellow in Robotics Institute of Carnegie Mellon University and National University of Singapore. From 2022, he is a New Hundred-Talent Program faculty, directing Grasp Lab at Zhejiang University, China. He is an associate editor of IEEE Robotics and Automation Letter, IEEE Transactions on Automation Science and Engineering, ICRA, IROS and AIM. His current research interests include robotic perception and manipulation in unstructured environments, robotic gripper/hand. 
% \end{IEEEbiography}

\end{document}